\documentclass[pmlr,twocolumn,10pt]{jmlr} 
\usepackage{verbatim}
\usepackage{attachfile}
\usepackage{float}
\usepackage{caption}
\usepackage{balance}
\usepackage{multirow}
\usepackage[english]{babel}
\usepackage{microtype}
\usepackage{lscape}

\usepackage{booktabs}

\usepackage{siunitx}
\usepackage{graphicx} 
\usepackage{hyperref}

\usepackage[switch]{lineno}

\theorembodyfont{\upshape}
\theoremheaderfont{\scshape}
\theorempostheader{:}
\theoremsep{\newline}

\jmlrvolume{287}
\jmlryear{2025}
\jmlrsubmitted{} 
\jmlrpublished{} 
\jmlrworkshop{Conference on Health, Inference, and Learning (CHIL) 2025}

\title{CaseReportBench: An LLM Benchmark Dataset for Dense Information Extraction in Clinical Case Reports}

\author{%
\Name{Xiao Yu Cindy Zhang} \Email{czhang@cmmt.ubc.ca} \\
\addr University of British Columbia
\AND
\Name{Carlos R. Ferreira} \Email{carlos.ferreira@nih.gov} \\
\addr National Institutes of Health 
\AND
\Name{Francis Rossignol} \Email{francis.rossignol@nih.gov} \\
\addr National Institutes of Health
\AND
\Name{Raymond T. Ng} \Email{rng@cs.ubc.ca } \\
\addr University of British Columbia
\AND
\Name{Wyeth Wasserman} \Email{wyeth@cmmt.ubc.ca}\\
\addr University of British Columbia
\AND
\Name{Jian Zhu} \Email{jian.zhu@ubc.ca}\\
\addr University of British Columbia
}

\begin{document}

\maketitle

\begin{abstract}
Rare diseases, including Inborn Errors of Metabolism (IEM), pose significant diagnostic challenges. Case reports serve as key but computationally underutilized resources to inform diagnosis. Clinical dense information extraction refers to organizing medical information into structured predefined categories. Large Language Models (LLMs) may enable scalable information extraction from case reports but are rarely evaluated for this task. We introduce \textbf{CaseReportBench}, an expert-annotated dataset for dense information extraction of case reports (focusing on IEMs). Using this dataset, we assess various models and promptings, introducing novel strategies of \textbf{category-specific prompting} and \textbf{subheading-filtered data integration}. Zero-shot chain-of-thought offers little advantage over zero-shot prompting. \textbf{Category-specific prompting} improves alignment to benchmark. Open-source \textbf{Qwen2.5:7B} outperforms \textbf{GPT-4o} for this task. Our clinician evaluations show that LLMs can extract clinically relevant details from case reports, supporting rare disease diagnosis and management. We also highlight areas for improvement, such as LLMs' limitations in recognizing negative findings for differential diagnosis. This work advances LLM-driven clinical NLP, paving the way for scalable medical AI applications.

\end{abstract}

\paragraph*{Data and Code Availability}
This study used de-identified case reports from the non-commercial PMC Open Access Subset (CC BY-NC, CC BY-NC-SA, CC BY-NC-ND), accessed via {PMC FTP} \footnote{\url{https://ftp.ncbi.nlm.nih.gov/pub/pmc/oa_bulk/oa_noncomm/xml/}} on February 3, 2024. \textbf{CaseReportBench} is publicly available on Hugging Face\footnote{\url{https://huggingface.co/datasets/cxyzhang/caseReportBench_ClinicalDenseExtraction_Benchmark}}, with source code on GitHub\footnote{\url{https://github.com/cindyzhangxy/CaseReportBench}}, under a CC BY-NC license.

\paragraph*{Institutional Review Board (IRB)}
IRB approval was not required, as only publicly available, de-identified data were used.

\section{Introduction}
\label{sec:intro}

Rare diseases encompass over 7,000 heterogeneous conditions that are often chronic, progressive, and debilitating \citep{chung2022rare}. Among them, \textbf{Inborn Errors of Metabolism (IEMs)} is a group of over 1450 genetic disorders that disrupt metabolic pathways and contribute to significant morbidity and mortality \citep{blau2014physician, ferreira2021international}. It affects approximately 1 in 2,500 births, posing a substantial global health burden \citep{applegarth2000incidence}.  Case reports serve as key clinical documentation for rare diseases, offering actionable insights when robust guidelines are lacking. \citep{carey2010importance}. They are essential for researching novel disorders and uncovering unexpected disease-symptom associations \citep{vandenbroucke2001defense}. 

Extracting dense, granular information into a structured format can be fundamental for developing reliable AI-based diagnostic aids. However, manual extraction of such information is often error-prone, and impractical in busy clinical settings. Case reports typically follow a standard format, including title, abstract, introduction, case presentation, discussion, conclusion, and references \citep{riley2017care}. Specifically, the case presentation section provides a detailed account of patient assessment, such as system reviews and history, essential for disease diagnosis. This study aims to establish a benchmark for evaluating dense information extraction from clinical case reports, focusing on medical conditions, and identifying effective LLM-based methods for this task. Our contributions are as follows:
\begin{itemize}
    \item We present \textbf{CaseReportBench}, an expert-crafted medical benchmark designed for dense information extraction from case reports, and evaluate open-access LLMs and GPT-4o on this task. 
    \item We evaluate diverse prompting and data integration strategies for dense information extraction. We propose novel techniques, including category-specific extraction prompting and a subheading-filtered approach, to enhance dense information extraction by LLMs.
    \item We conducted a clinical evaluation on LLM-extracted outputs, demonstrating their potential to reduce manual extraction efforts while identifying challenges.
\end{itemize}

\section{Related Work}

Automating dense information extraction presents several challenges. Clinical text is often information-packed, complicating systematic extraction \citep{landolsi2023information, wiest2024llm}. Extracting meaningful clinical insights requires linking multiple isolated entities to capture the full context \citep{steinkamp2019toward}. Furthermore, an adaptable approach is required to accommodate the variety and complexity of information types—such as patient history, laboratory findings, and imaging \citep{steinkamp2019toward}. 

Various benchmark datasets exist for specific medical NLP tasks. Datasets such as BIOSSES, containing sentence pairs from biomedical literature \citep{souganciouglu2017biosses}, and MedSTS, derived from EHRs \citep{wang2020medsts}, both serve as key benchmarks for medical sentence similarities. BioCreative V Chemical Disease Relation corpus, which annotates chemical-disease interactions from PubMed articles \citep{li2016biocreative}, was used for named entity recognition. PMC-Patients dataset, which consists of patient summaries extracted from PubMed Central (PMC) \citep{zhao2022pmc}, provides a benchmark for a retrieval-based clinical decision support system. MedNLI, a dataset based on MIMIC-III, has been widely used for evaluating clinical natural language inference \citep{romanov2018lessons}. 

While existing benchmarks provide valuable insights, they primarily focus on isolated NLP tasks. In contrast, dense information extraction requires capturing structured knowledge across multiple dimensions. Relationships extracted in existing datasets (e.g., PMC-Patients \citep{zhao2022pmc}) are often binary and insufficient for identifying multiple relations, attributes, and temporal aspects required by dense information extraction.  Moreover, widely used benchmarks (e.g., MIMIC) \citep{johnson2016mimic}, focus on single-center EHR, limiting their applicability to diverse case report narratives. Thus, benchmarks specifically designed for dense information extraction from non-EHR clinical data remain largely unexplored. Our study seeks to overcome the limitations of existing benchmarks in dense information extraction.  

LLMs have proven effective in extracting fine-grained medical information from clinical narratives \citep{fornasiere2024medical, goel2023llms, agrawal2022large}. This study evaluates their capability for dense information extraction from medical case reports by testing various models, prompting strategies, and data integration methods. Additionally, we explore evaluation metrics for dense information extraction.

\section{Dataset Construction}
\begin{figure*}[t]
    \centering
    \includegraphics[width=0.8\textwidth]{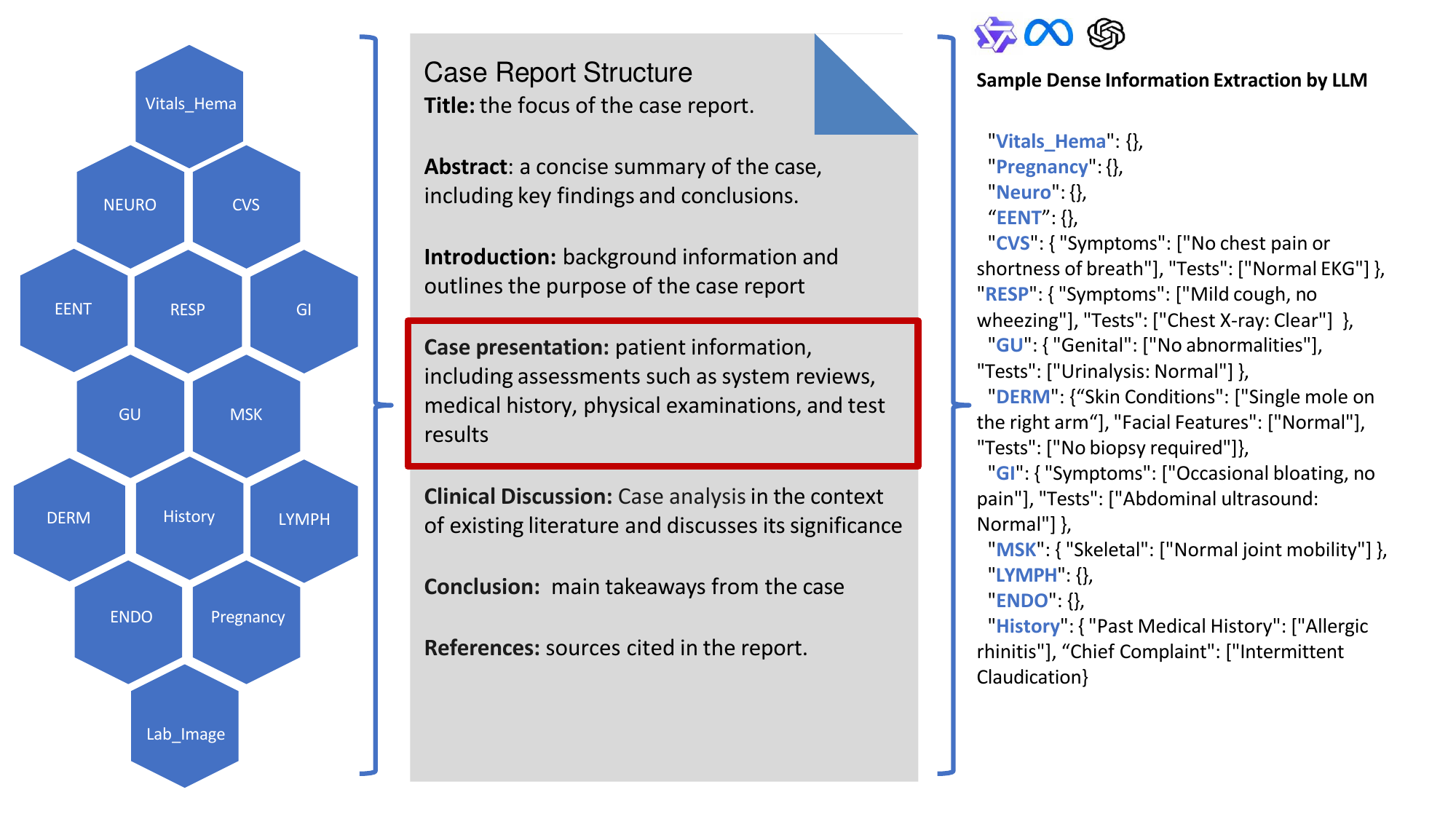}
    \caption{ Overview of CaseReportBench: Structuring Clinical Case Reports for Dense Information Extraction. This figure illustrates the systematic organization of case reports, the 14 key clinical categories used for structured extraction, and sample outputs from LLM-based extraction. }
    \label{fig:categories}
\end{figure*}

Our goal is to extract clinically relevant insights from case reports, converting information-dense narratives into a structured format that aligns with real-world clinical assessments. To this end, We focused on 14 key clinical categories: Vitals and Hematology Findings (\textbf{Vitals\_Hema}), Eyes, Ears, Nose, and Throat (\textbf{EENT}), Neurology (\textbf{NEURO}), Cardiovascular System (\textbf{CVS}), Respiratory System (\textbf{RESP}), Gastrointestinal System (\textbf{GI}), Genitourinary System (\textbf{GU}), Musculoskeletal System (\textbf{MSK}), Dermatology (\textbf{DERM}), Lymphatic System (\textbf{LYMPH}), Endocrinology (\textbf{ENDO}), \textbf{Pregnancy}, Laboratory and Imaging (\textbf{Lab\_Image}), and \textbf{History}, as shown in Figure ~\ref{fig:categories}, adapted from the standard Inpatient Work-Up and Monitoring Form\footnote{\url{https://blogs.ubc.ca/oeetoolbox/2019/02/patient-work-up-from-sample-template-inpatient/}}.

\subsection{Dataset Preprocessing}
We adopted a privacy-conscious approach to derive the dataset from open-access case reports in PMC. Since case reports are often lengthy and contain subsections irrelevant to patient assessments, we developed a tailored preprocessing framework to filter case report sections. Specifically, case reports are divided into subsections based on their corresponding subheadings. We used regular expressions to classify subsections under subheadings containing predefined word roots and to exclude irrelevant sections, such as those titled ``References" or ``Disclaimers". Additionally, we implemented a predefined exclusion list to filter out articles whose titles indicate a focus on treatments or procedures rather than medical conditions. The exclusion criteria and filtered section list are available in the supplementary materials.

The framework captures terminology variations within each medical category, prioritizing commonly used clinical terms. While not exhaustive, the list is openly available to support ongoing expansion and refinement.

\subsection{Case Report Selection}
Following the preprocessing of case reports, we utilized \textbf{IEMbase}\footnote{\url{https://www.iembase.org/}}, a knowledge base for IEMs to capture a rare disease-focused case report dataset. IEMbase encompasses an expert-curated collection of 1,907 disorders with robust coverage of confirmed IEMs and disorders with overlapping clinical presentations. From IEMbase we extracted disease names, acronyms, abbreviations, and aliases, and used these terms to identify case reports for inclusion.

\begin{figure*}[t]
\centering
    \includegraphics[width=0.8\textwidth]{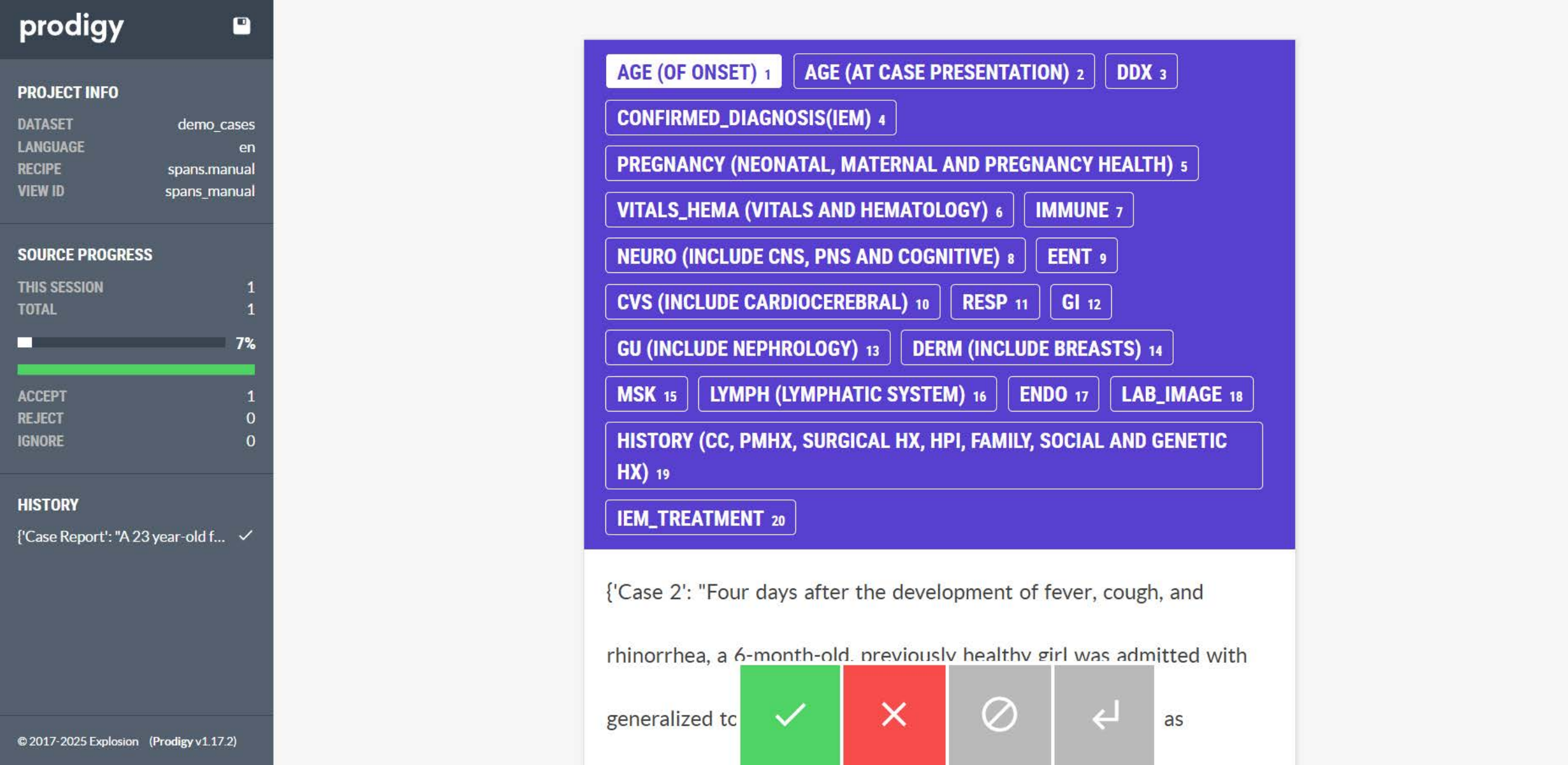}
    \caption{The Prodigy SPAN annotation interface for dense information extraction.}

    \label{appendix:prodigy_interface}
\end{figure*}

\subsection{Expert-Guided Annotation and Dataset Reconciliation}

The annotation and dataset reconciliation process consisted of multiple steps for developing this expert-annotated dataset (Appendix~\ref{appendix:annotation_workflow}). The first phase involves the first author and two specialists collaboratively designing category labels and developing annotation guidelines, provided in the supplementary material. These guidelines were reviewed with the annotators to ensure clarity and reduce ambiguity. Following this, these specialists independently performed information extraction on the dataset using \textit{Prodigy} annotation tool (Fig. \ref{appendix:prodigy_interface}) \citep{Prodigy:2018}

\subsection{Inter-Annotator Agreement and Token Set Ratio}

Inter-Annotator Agreement (IAA) is crucial in annotation tasks, but traditional metrics like Krippendorff’s $\alpha$ are challenging to interpret due to task-dependent thresholds and varying distance functions \citep{braylan2022measuring}. Dense information extraction poses additional challenges due to variability in extracted spans caused by subjective clinical classifications, often leading to low agreement scores even when there is no true disagreement (e.g., \textit{``progressive muscle weakness''} vs. \textit{``muscle weakness over 5 years''}). To find the right metric for IAA evaluation, we explored several string-based evaluation metrics, more suitable for dense information extraction tasks. 
 
\textbf{Levenshtein Similarity (Levenshtein)}  quantifies text similarity based on the Levenshtein distance, which counts the minimum character edits (insertions, deletions, substitutions) needed to transform one string into the reference. It is scored from 0 to 100, with higher values indicating greater similarity between the extracted and reference text. \textbf{Exact Match} measures the percentage of instances where the model’s output exactly matches the reference text, where a score of 100 indicates a perfect match and 0 indicates no match. \textbf{Token Set Ratio (TSR(\%))}, which measures how effectively the model selects the correct tokens from the reference text, is scored from 0 to 100, where a higher score indicates better token-level alignment. It is computed using the FuzzyWuzzy library\footnote{\url{https://github.com/seatgeek/fuzzywuzzy}} for aligning two dense extractions. Given token sets $T_1$ and $T_2$ from two extractions, TSR first computes their intersection $I = T_1 \cap T_2$ and differences ($D_1 = T_1 \setminus T_2$, $D_2 = T_2 \setminus T_1$). These sets are then transformed into strings, and similarity is computed using Levenshtein distance. The final TSR is determined as the maximum similarity score across all pairwise comparisons.

Since the three metrics have shown consistent results  (Appendix~\ref{tab:annotation_metrics}), the author with a clinical credential and current NLP training performed an iterative review followed by discussions with the two specialists to resolve discrepancies for instances with TSR(\%)$<$ 30. During this process, the annotation guidelines were refined and clarified to address ambiguities that emerged in the initial phase. This iterative feedback loop aimed to ensure that evolving insights can be incorporated into the guidelines for improving consistency and accuracy.
\begin{figure*}[h]
\centering
    \includegraphics[width=\textwidth]{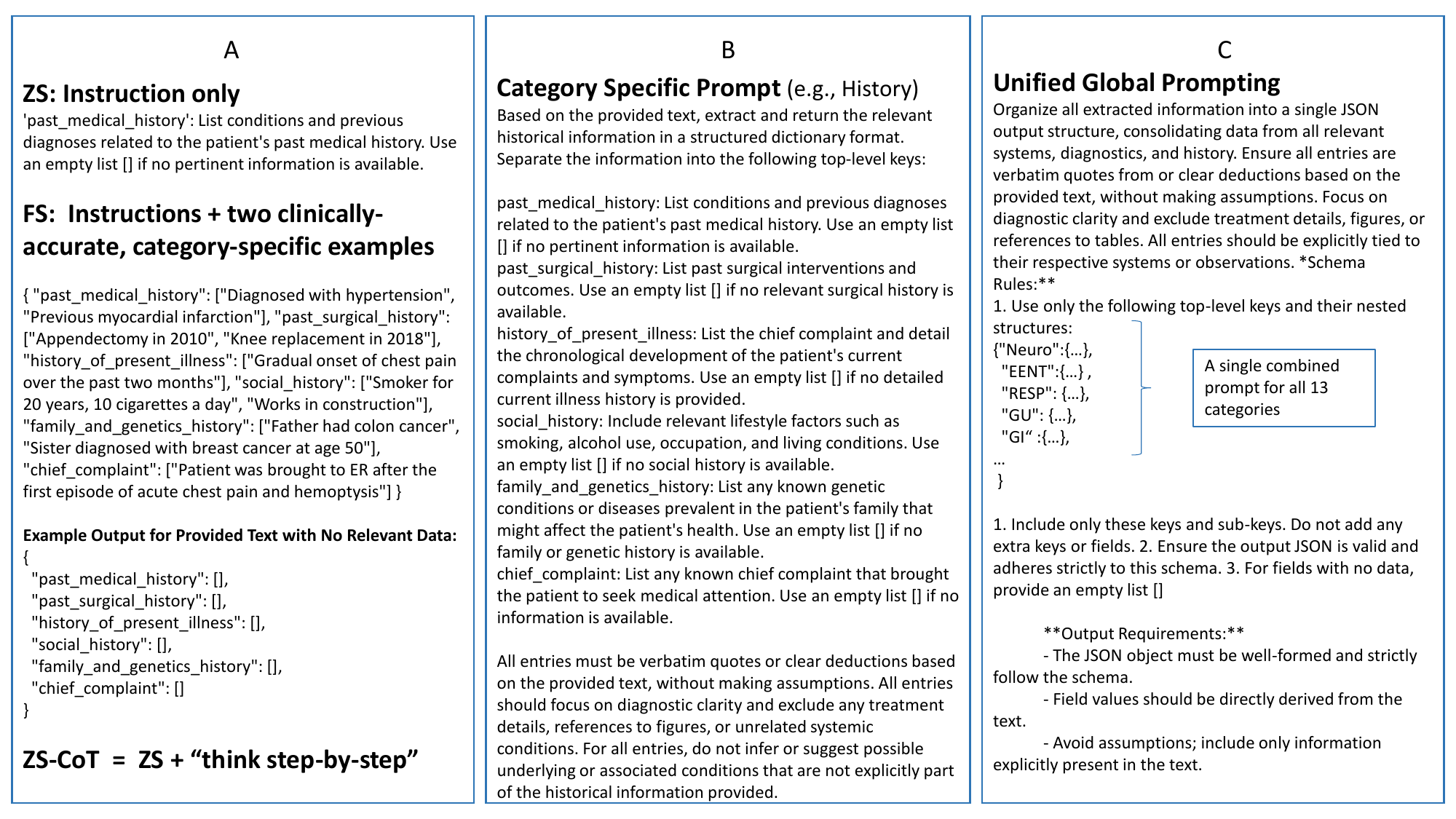}
    \caption{Comparison of different prompting strategies: (A) Examples of ZS, FS, and ZS-CoT promptings; (B) a category-specific prompt demonstrated using the ``History'' category as an example and (C) a unified prompt applied across all categories instead of individual category-specific prompts }
    \label{image:prompting}
    
\end{figure*}

\subsection{Expert Annotation Concordance and Finalized Dataset Analysis}
The initial comparison of domain expert annotations in the gold standard for dense information extraction reveals variation in annotation proportions across certain categories, such as EENT, suggesting differences in interpretation or emphasis among clinicians. Despite these discrepancies, other categories, such as Pregnancy, exhibit high inter-annotator agreement, indicating strong alignment in experts' understanding and application of the annotation guidelines (Appendix~\ref{tab:annotation_metrics} and \ref{appendix:iaa}).  422 instances with TSR(\%)$<$ 30 were identified for further reconciliation, primarily due to differences in guideline interpretation by each specialist. Key disagreements occurred in categories such as Lab\_Image, History, and MSK, where, for instance, ECG imaging was categorized as CVS rather than Lab\_Image. Structural and functional observations in MSK and GU also led to variations.
        
The final dataset contains 138 case report entries with a primary rare disease focus. Table \ref{tab:annotation_string_length} presents the annotation coverage and average string length for different categories in the \textbf{CaseReportBench} dataset. The \textbf{Annotation (\%)} shows the proportion of the 138 case reports in \textbf{CaseReportBench} that contain at least one annotation for each respective category. For instance, Lab\_Image and History have the highest annotation coverage at 98.55\%, meaning that 136 out of 138 cases include annotations for these categories. In contrast, LYMPH has the lowest coverage at 1.45\%. The \textbf{Average String Length} by word count represents the average length of annotated text spans across all cases within each category. \textbf{Lab\_Image} exhibits the longest average annotation length at 270.79 words, indicating that laboratory and imaging-related annotations tend to be more extensive. \textbf{Pregnancy} and \textbf{History} also show relatively long annotations, averaging 144.10 words and 204.95 words, respectively. On the other hand, \textbf{GI} and \textbf{LYMPH} categories have the shortest annotations, averaging 37.23 words and 47.50 words, respectively.

\begin{table}
    \centering
   
    \resizebox{0.5\textwidth}{!}{ 
    \begin{tabular}{lcc}
        \hline
        \textbf{Category} & \textbf{Annotation (\%)} & \textbf{Average String Length}\\
        \hline
        Vitals\_Hema  & 59.42  & 80.98   \\
        Neuro         & 68.84  & 103.59  \\
        EENT          & 49.28  & 84.92   \\
        CVS           & 35.51  & 84.36   \\
        RESP          & 17.39  & 54.56   \\
        GI            & 45.65  & 37.23   \\
        GU            & 41.30  & 87.05   \\
        MSK           & 47.83  & 82.92   \\
        DERM          & 49.28  & 111.16  \\
        Lab\_Image    & 98.55  & 270.79  \\
        LYMPH         & 1.45   & 47.50   \\
        History       & 98.55  & 204.95  \\
        ENDO          & 28.99  & 91.87   \\
        Pregnancy     & 31.16  & 144.10  \\
        \hline
    \end{tabular}
    }
     \caption{CaseReportBench annotation coverage and average string length by category.}

    \label{tab:annotation_string_length}
\end{table}

\section{Benchmarking Dense Medical Information Extraction}

We evaluated five LLMs for dense medical information extraction: \textbf{Qwen2:7B-Instruct} \citep{hui2024qwen2}, \textbf{Qwen2.5:7B-Instruct}, \textbf{Qwen2.5:32B-Instruct} \citep{yang2024qwen2}, \textbf{Llama3:8B-Instruct} \citep{dubey2024llama}, and \textbf{GPT-4o} \citep{islam2024gpt}. 

\subsection{Filtered Subheading Data Integration}
To enhance the precision of extracted information, we investigated an approach that aligns category-specific prompts with relevant text sections. In this method, subheadings were filtered using a predefined list ( Supplementary) to match prompts with appropriate content. For example, the above ``History"-specific prompt will be applied to sections with the subheading ``Patient History", rather than ``Laboratory Findings" to ensure that information extraction remains contextually appropriate. For subheadings that cannot be classified into a specific category (e.g. ``Patient Assessment"), each category-specific prompt will be applied to the text for dense information extraction.

\subsection{Category-Specific Prompting}
Previous studies have used broad, uniform prompts applied to the entire input text \citep{richter2024clinical, sivarajkumar2023healthprompt}, potentially overlooking nuances across different information categories. To address this, we introduce \textbf{category-specific prompting}, where medically relevant prompts were crafted for each category by the author with a clinical credential and current NLP training.  Figure ~\ref{image:prompting}B) illustrates a ZS variant using the ``History" category as an example. The LLM was instructed to extract information under key subheadings, including past medical history, history of present illness, social history, family and genetic history, and chief complaint, ensuring a structured extraction process that aligns with real-world patient assessments. In contrast, Figure ~\ref{image:prompting}A) demonstrates the FS variant, where the researcher provided medically accurate examples specific to each category, guiding the model with correct extractions for relevant data and an empty output format when no relevant information was present. This approach enhances alignment with clinical reasoning while minimizing irrelevant outputs. 
To leverage category-specific knowledge at varying levels of granularity, we explored three prompting strategies:

\subsubsection{Filtered Category-Specific Prompting (FCSP)}
FCSP combined structured data filtering with tailored prompts for predefined medical categories. Each category was assigned a distinct, individually crafted prompt to enhance extraction accuracy. For instance, a history-specific prompt was designed to extract pertinent historical details, as illustrated in Figure \ref{image:prompting}B. Each case report was segmented into sections based on subheadings and mapped to predefined categories. The category-specific prompt was then applied to a text segment assigned to the corresponding medical category.  

\subsubsection{Uniform Category-Specific Prompting (UCP)}
Unlike FCSP, which applied category-specific prompts to corresponding filtered text segments, UCP applied prompts to the entire case report text without prior filtering. This approach allows the model to infer and extract relevant information without relying on subheading-based segmentation. Additionally, it eliminates the need for preprocessing and crafted filters, which require medical knowledge and an understanding of the text structure.

\subsubsection{Uniform Global Prompting (UGP)}
Both FCSP and UCP require multiple inference calls to the LLM, which can be computationally intensive. As a baseline, we also implemented UGP. UGP consolidated all predefined medical categories into a single, unified prompt, as illustrated in Figure \ref{image:prompting}C. Rather than applying category-specific prompts individually, this method instructed the LLM to extract information across multiple categories simultaneously in a single prompt. We hypothesize that this approach may further enhance efficiency by reducing the need for multiple sequential prompt applications on each text segment, offering a more scalable solution for large-scale automated extraction tasks.

\subsection{Prompting Strategies Across Methods} The above data integration methods were implemented in conjunction with the prompting strategies: ZS and FS. In addition,  ZS-CoT was used in combination with FCSP and UCP for Qwen2.5:32B-Instruct.

\subsection{Implementation}
Experiments for all open-access models were conducted using the \cite{ollama} framework on an NVIDIA Tesla V100 GPU. All open-access models leveraged 4-bit quantization for enhanced memory efficiency during inference. For GPT-4o, we used the proprietary API provided by OpenAI.

\section{Results and Discussion}
\begin{table*}[t]
\centering
\resizebox{0.8\textwidth}{!}{
\begin{tabular}{lllrrr}
\toprule
 \textbf{Model : Parameters} &  \textbf{Prompting} &  \textbf{Method} &  \textbf{TSR (\%)↑} &  \textbf{Levenshtein ↑} &  \textbf{ExactMatch ↑}\\
\midrule
qwen2:7b & FS & UGP & 25.107 & 0.043 & 0.001 \\
qwen2:7b & ZS & UGP & 27.748 & 0.057 & 0.001 \\
qwen2:7b & ZS & FCSP & 43.477 & 0.297 & 0.211 \\
qwen2:7b & ZS & UCP & 43.513 & 0.298 & 0.210 \\
qwen2:7b & FS & UCP & 48.592 & 0.343 & 0.259 \\
\textbf{qwen2:7b} & \textbf{FS} & \textbf{FCSP} & \textbf{48.693} & \textbf{0.344} & \textbf{0.260} \\
\midrule
qwen2.5:7b & ZS & UGP & 28.432 & 0.071 & 0.001 \\
qwen2.5:7b & FS & UGP & 28.627 & 0.085 & 0.031 \\
qwen2.5:7b & ZS & FCSP & 51.946 & 0.423 & 0.345 \\
qwen2.5:7b & ZS & UCP & 52.041 & 0.425 & 0.345 \\
qwen2.5:7b & FS & FCSP & 56.377 & 0.453 & 0.383 \\
\textbf{qwen2.5:7b} & \textbf{FS} & \textbf{UCP} & \textbf{56.456} & \textbf{0.454} & \textbf{0.383} \\
\midrule
qwen2.5:32b & FS & UGP & 23.378 & 0.032 & 0.001 \\
qwen2.5:32b & ZS & UGP & 28.805 & 0.082 & 0.001 \\
qwen2.5:32b & ZSCOT & UCP & 49.680 & 0.385 & 0.284 \\
qwen2.5:32b & ZS & UCP & 49.917 & 0.386 & 0.285 \\
qwen2.5:32b & ZS & FCSP & 50.203 & 0.387 & 0.288 \\
qwen2.5:32b & ZSCOT & FCSP & 50.203 & 0.387 & 0.288 \\
qwen2.5:32b & FS & UCP & 51.332 & 0.387 & 0.292 \\
\textbf{qwen2.5:32b} & \textbf{FS} & \textbf{FCSP} & \textbf{51.467} & \textbf{0.388} & \textbf{0.296} \\
\midrule
llama3:8b & FS & UGP & 26.072 & 0.047 & 0.001 \\
llama3:8b & ZS & UGP & 28.732 & 0.066 & 0.001 \\
llama3:8b & ZS & FCSP & 49.462 & 0.343 & 0.259 \\
llama3:8b & ZS & UCP & 49.536 & 0.344 & 0.258 \\
llama3:8b & FS & UCP & 50.255 & 0.362 & 0.283 \\
\textbf{llama3:8b} & \textbf{FS} & \textbf{FCSP} & \textbf{50.351} & \textbf{0.363} & \textbf{0.285} \\
\midrule
gpt4o & FS & UGP & 23.416 & 0.033 & 0.001 \\
gpt4o & ZS & UGP & 23.452 & 0.033 & 0.001 \\
gpt4o & FS & UCP & 30.862 & 0.136 & 0.046 \\
gpt4o & FS & FCSP & 31.051 & 0.132 & 0.044 \\
gpt4o & ZS & FCSP & 31.107 & 0.132 & 0.044 \\
\textbf{gpt4o} & \textbf{ZS} & \textbf{UCP} & \textbf{31.148} & \textbf{0.135} & \textbf{0.046} \\
\bottomrule
\end{tabular}}
\caption{Performance of models under different promptings and data integration methods on string-level metrics. ↑ indicates the higher the better.}
\label{tab:str_metrics}
\end{table*}

\begin{table*}[h!]
\centering

\resizebox{0.8\textwidth}{!}{ 
\begin{tabular}{lrrrr}
\toprule
 & \textbf{BLEU-1 Avg ↑} & \textbf{BLEU-4 Avg ↑} & \textbf{ROUGE-L Avg ↑} & \textbf{Hallucination (\%)↓} \\
\midrule
qwen2:7b\_FS\_UGP & 0.011 & 0.000 & 0.008 & 92.860 \\
qwen2:7b\_ZS\_UGP & 0.017 & 0.000 & 0.011 & 92.860 \\
qwen2:7b\_ZS\_UCP & 0.280 & 0.016 & 0.252 & 64.740 \\
qwen2:7b\_ZS\_FCSP & 0.281 & 0.016 & 0.252 & 64.680 \\
qwen2:7b\_FS\_UCP & 0.324 & 0.012 & 0.299 & 57.510 \\
\textbf{qwen2:7b\_FS\_FCSP} & \textbf{0.326} & \textbf{0.012} & \textbf{0.299} & \textbf{57.320} \\

\midrule
qwen2.5:7b\_ZS\_UGP & 0.018 & 0.000 & 0.012 & 92.860 \\
qwen2.5:7b\_FS\_UGP & 0.047 & 0.004 & 0.038 & 88.240 \\
qwen2.5:7b\_ZS\_UCP & 0.420 & 0.025 & 0.392 & 44.110 \\
qwen2.5:7b\_ZS\_FCSP & 0.421 & 0.026 & 0.391 & 44.210 \\
qwen2.5:7b\_FS\_UCP & 0.462 & 0.024 & 0.427 & 38.550 \\
\textbf{qwen2.5:7b\_FS\_FCSP} & \textbf{0.463} & \textbf{0.025} & \textbf{0.426} & \textbf{38.550} \\

\midrule
qwen2.5\_32b\_FS\_UGP & 0.002 & 0.000 & 0.001 & 92.860 \\
qwen2.5\_32b\_ZS\_UGP & 0.018 & 0.000 & 0.012 & 92.860 \\
qwen2.5\_32b\_ZS\_UCP & 0.371 & 0.030 & 0.343 & 55.410 \\
qwen2.5\_32b\_ZSCOT\_UCP & 0.371 & 0.031 & 0.343 & 55.570 \\
qwen2.5\_32b\_ZS\_FCSP & 0.376 & 0.031 & 0.346 & 54.960 \\
qwen2.5\_32b\_ZSCOT\_FCSP & 0.376 & 0.031 & 0.346 & 54.960 \\
qwen2.5\_32b\_FS\_UCP & 0.383 & 0.033 & 0.348 & 54.330 \\
\textbf{qwen2.5\_32b\_FS\_FCSP} & \textbf{0.386} & \textbf{0.033} & \textbf{0.350} & \textbf{53.870} \\

\midrule
llama3:8b\_FS\_UGP & 0.010 & 0.000 & 0.007 & 92.860 \\
llama3:8b\_ZS\_UGP & 0.018 & 0.000 & 0.012 & 92.860 \\
llama3:8b\_ZS\_UCP & 0.338 & 0.020 & 0.303 & 61.160 \\
llama3:8b\_ZS\_FCSP & 0.339 & 0.021 & 0.303 & 61.110 \\
llama3:8b\_FS\_UCP & 0.365 & 0.024 & 0.329 & 53.230 \\
\textbf{llama3:8b\_FS\_FCSP} & \textbf{0.367} & \textbf{0.024} & \textbf{0.331} & \textbf{52.970} \\

\midrule
gpt4o\_ZS\_UGP & 0.003 & 0.000 & 0.001 & 92.860 \\
gpt4o\_FS\_UGP & 0.004 & 0.000 & 0.001 & 92.860 \\
gpt4o\_ZS\_FCSP & 0.091 & 0.007 & 0.077 & 85.160 \\
gpt4o\_FS\_FCSP & 0.092 & 0.007 & 0.078 & 85.160 \\
gpt4o\_ZS\_UCP & 0.093 & 0.007 & 0.079 & 84.880 \\
\textbf{gpt4o\_FS\_UCP} & \textbf{0.093} & \textbf{0.007} & \textbf{0.080} & \textbf{84.880} \\
\bottomrule
\end{tabular}
}
\caption{Performance of models under different promptings and data integration methods on traditional NLP metrics. ↑ indicates the higher the better. }
\label{tab:token_level_metrics}
\end{table*}

\paragraph{Qwen2.5:7B Outperforms all Tested Models for Dense Information Extraction.}

Across all models, we evaluated hallucinations—defined as the percentage of LLM-extracted information not present in the benchmark set—alongside string-based metrics (TSR(\%), EM) and traditional metrics (BLEU, ROUGE).

Among the models, \textsc{Qwen2.5-7B} demonstrated the highest alignment with expert annotations for dense information extraction in medical case reports. This was followed by \textbf{Llama3-8B} and \textbf{Qwen2.5-32B}, with \textbf{Qwen2-7B} performing slightly worse. In contrast, \textbf{GPT-4o} exhibited the lowest performance (Table~\ref{tab:str_metrics}).Using traditional evaluation metrics, \textbf{Qwen2.5-7B} consistently achieved \textbf{high BLEU and ROUGE scores} while maintaining \textbf{low hallucination rates}. Conversely, \textbf{GPT-4o} showed the highest hallucination levels (Table~\ref{tab:token_level_metrics}).

\paragraph{FCSP Achieves Comparable Performance with Better Efficiency; UGP Largely Fails to Align with Prompt.}

The average TSR(\%) scores for \textbf{FCSP} and \textbf{UCP} were similar, likely due to the dataset’s predominance of single-subheading sections, limiting the benefit of subheading-based filtering. However, \textbf{FCSP} was more efficient, reducing computation time by \textbf{6.2\%}, as \textbf{UCP} applied prompts indiscriminately to all paragraphs, increasing overhead—particularly problematic for longer texts.

In contrast, \textbf{UGP} consistently failed to follow prompt instructions, generating outputs like \{`error': [], `Lab\_Image': []\}, which were not comparable to the \textbf{CaseReportBench} using TSR. \textbf{UGP} consistently demonstrated the weakest performance, as indicated by \textbf{low BLEU and ROUGE-L scores}, reflecting poor alignment with the reference text and high hallucination rates. Lastly, while \textbf{Qwen2.5} and \textbf{GPT\-4o} extracted additional details beyond the predefined prompt instructions, this also introduced alignment challenges with the evaluation format. This suggests that while these models demonstrate higher sensitivity to information, they demand different metrics for fair evaluation. 

\paragraph{Open-Access Models Outperform GPT-4o, and Larger Model Sizes Do Not Always Improve Dense Extraction.}
We observed that \textbf{open-access LLMs} adhere to prompt instructions more faithfully than \textbf{GPT-4o}, which struggles with verbatim extraction, merges details, introduces inferences and deviates from structured formatting. These limitations suggest that \textbf{GPT-4o}'s alignment with structured medical data extraction is weaker than expected, emphasizing the importance of instruction fidelity in high-stakes domains like healthcare. 

Additionally, larger models do not guarantee superior performance—\textbf{Qwen2.5:7B} outperforms \textbf{Qwen2.5:32B} despite the latter's higher parameter count, and \textbf{GPT-4o} underperforms relative to all evaluated models, despite its presumed advantage in training data volume. This indicates that architecture, instruction adherence, and formatting compliance are more critical than sheer model size in dense medical information extraction. The lower performance of GPT-4o is likely due to the \textbf{stringent evaluation metrics} used, which penalize minor variations (e.g., synonyms, rewording). While GPT-4o excels in \textbf{fluency and general reasoning}, its broad training on vast datasets increases the risk of \textbf{spurious associations} and \textbf{hallucinations}, particularly in rare medical cases. In high-stakes healthcare applications, textbf{factual precision outweighs creativity}, and open-source models—being more adaptable and controllable—may align better with strict recall-based evaluations

When comparing the performance of \textbf{Qwen2.5:32B} in the \textbf{FCSP} and \textbf{UCP} settings, we found \textbf{ZS} and \textbf{ZS-CoT} were inferior to simply FS prompting. This suggests that step-by-step reasoning by \textbf{CoT}, \textbf{does not confer additional benefits} for dense information retrieval tasks. These findings underscore that task-specific alignment, instruction fidelity, and optimal prompting strategies may be key determinants of performance in medical NLP applications.

Since only minor differences were observed between standard \textbf{ZS} and \textbf{ZS-CoT}, \textbf{ZS-CoT} was only applied to \textbf{Qwen2.5:32B} without extending to other models. We acknowledged that \textbf{FS\-CoT} was not pursued in this study due to its higher computational cost and anticipated limited improvements over FS prompting, given the structured nature of the extraction task. Exploring ZS\-CoT on different models and crafting clinically accurate reasoning step examples for FS\-CoT represents a future direction of the study.

\begin{figure*}[t]
    \includegraphics[width=\textwidth]{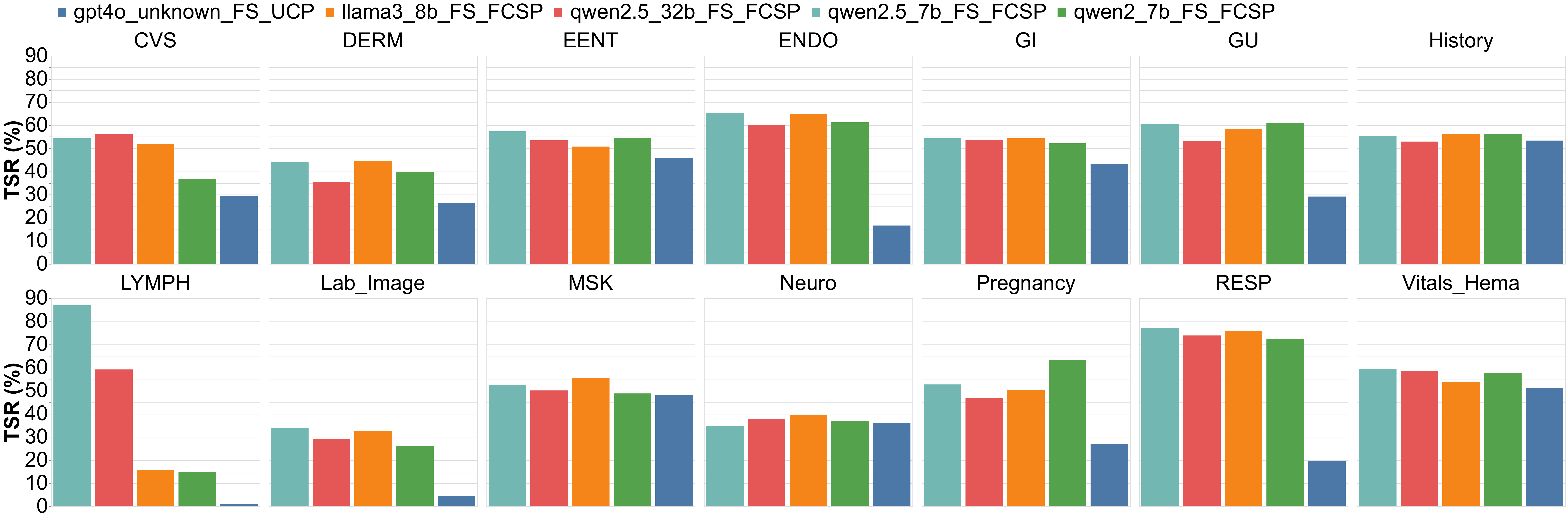}
\caption{Per-category TSR(\%) across the best-performing prompting strategies in each model}
    \label{image:per_cat_tsr}
    
\end{figure*}

\paragraph{Open-Source Models and GPT-4o Shows Different Per-Category Extraction Profile.}

Open-source models achieved the highest extraction performance in several categories, such as History, EENT, and Vitals\_Hema, likely due to the presence of distinctive medical terminologies in these domains. \textbf{GPT-4o} demonstrated consistent performance across most categories, except for \textbf{RESP}, where it exhibited a notable decline. Both open-source models and \textbf{GPT-4o} showed equally low performance in the Lab\_Image category, likely due to its inherent complexity, which includes diverse numerical values, abbreviations, and intricate medical terminology.

Figure \ref{image:per_cat_tsr} presents the mean TSR(\%) across different categories for the best-performing strategy in each model. Compared to open-access models, \textbf{GPT-4o}'s performance was consistently lower across all categories. However, it achieved comparable performance in the \textbf{History} and \textbf{ENDO} categories, possibly due to its ability to extract more granular details, particularly in sections that contain fine-grained details, like ENDO.  

It is important to note that \textbf{LYMPH} content was underrepresented in the expert annotation, making it insufficient to draw conclusive insights regarding its extraction performance.

\begin{table*}[h]
\centering
\resizebox{0.8\textwidth}{!}{  
\begin{tabular}{@{}>{\raggedright\arraybackslash}p{4.6cm}p{5.4cm}c c c@{}}
\toprule
\textbf{Evaluation Criteria} & \textbf{Criteria} & \textbf{Physician 1} & \textbf{Physician 2} & \textbf{Average} \\ \midrule

\multirow{2}{*}{\textbf{1. Clinical Relevance↑}} 
& Accuracy of Key Information & 4 & 4 & 4 \\
& Critical Omissions & 4 & 4 & 4 \\ \midrule

\multirow{2}{*}{\textbf{2. Comprehensibility↑}} 
& Readability & 4 & 5 & 4.5 \\
& Conciseness & 5 & 5 & 5 \\ \midrule

\multirow{2}{*}{\textbf{3. Clinical Usability↑ }}
& Practicality & 4 & 4 & 4 \\
& Actionability & 3 & 3 & 3 \\ \midrule

\multirow{2}{*}{\shortstack[l]{\textbf{4. Error Impact}\\\textbf{Assessment↓}}}
& Severity of Errors & 2 & 2 & 2 \\
& Tolerance for Hallucination & 5 & 4 & 4.5 \\ \midrule

\multirow{2}{*}{\shortstack[l]{\textbf{5. Alignment with}\\\textbf{Clinical Judgment↑}}}
& Trustworthiness & 4 & 4 & 4 \\
& Contextual Appropriateness & 4 & 4 & 4 \\ \midrule

\textbf{6. Consistency↑} 
& Model Consistency Across Cases & 4 & 3 & 3.5 \\ \midrule

\textbf{7. Preference Scoring↑} 
& Overall Performance & 4 & 4 & 4 \\ 

\bottomrule
\end{tabular}
}

\caption{Consensus clinical evaluations of dense information extraction outputs from FS-FCSP by Llama3:8b. ↓ indicates the lower the better, and ↑ indicates the higher the better. }
\label{tab:evaluation_criteria}
\end{table*}

\subsection{Case Study: Clinician Assessment of LLM-Extracted Case Reports}

As a case study, we applied \textbf{FS-FCSP} with \textbf{LLaMA3:8B} for large-scale dense information extraction from case reports. To assess consistency, five LLaMA3-extracted cases (from the original 138 in \textbf{CaseReportBench}) were manually re-evaluated by the same clinicians post-dataset construction. This targeted evaluation focused on clinical relevance, usability, error impact, alignment with clinical judgment, and consistency—each rated on a 1–5 Likert scale (Supplementary), along with qualitative feedback for deeper insight.

Clinician evaluation for LLM on dense medical information extraction from the original LLama3 \textbf{CaseReportBench} was shown in Table \ref{tab:evaluation_criteria}. Both evaluators rated the model highly for its readability and conciseness, noting that its outputs were clear, easy to understand, and more concise than expert annotations. Practicality was also rated positively, with evaluators agreeing that the model's outputs could be integrated into clinical workflows. However, limitations in handling complex cases and categorization errors slightly impacted usability.

Key improvements included consistency across cases, contextual relevance, and handling omissions (e.g., negative findings). Evaluators noted strong trustworthiness, low clinically impactful errors, and good hallucination tolerance. Despite lower actionability scores, the model shows promise for streamlining workflows and reducing documentation effort.

\section{Conclusions}

This work introduces \textbf{CaseReportBench}, an expert-curated benchmark dataset for dense information extraction from clinical case reports. Covering 14 clinically relevant categories across 138 reports, it is the first benchmark to systematically evaluate LLM performance in structured clinical information extraction. We evaluated five LLMs using three data integration methods (FCSP, UCP, UGP) and three prompting strategies (ZS, FS, ZS-CoT). Our findings show that category-specific prompts (e.g., FSCSP, UCP) consistently outperform generic prompting (UGP), leading to more accurate and structured extractions. These results challenge the assumption that larger training data guarantees superior performance. Instead, task-specific instruction tuning appears more important than model size in high-stakes applications. Clinician evaluations further highlight the need for expert oversight: high token-level accuracy does not always translate to clinically actionable outputs. Nonetheless, LLM-assisted extraction reduced annotation time by approximately 24 hours across 138 cases, indicating efficiency gains and workflow support using LLM. \textbf{CaseReportBench} offers a practical foundation for clinically grounded LLM research and supports future expansion to broader disease domains.

\acks{We gratefully acknowledge support from the BC Children’s Hospital Research Institute (BCCHR) and computational resources provided by UBC Advanced Research Computing (Sockeye) and Digital Alliance of Canada.}

\bibliography{chil-sample}

\clearpage
\appendix

\onecolumn
\section{Expert Annotation and Consolidation Pipeline}
 \label{appendix:annotation_workflow}
 
\begin{figure*}[h]
    \centering
    \includegraphics[width=0.8\textwidth]{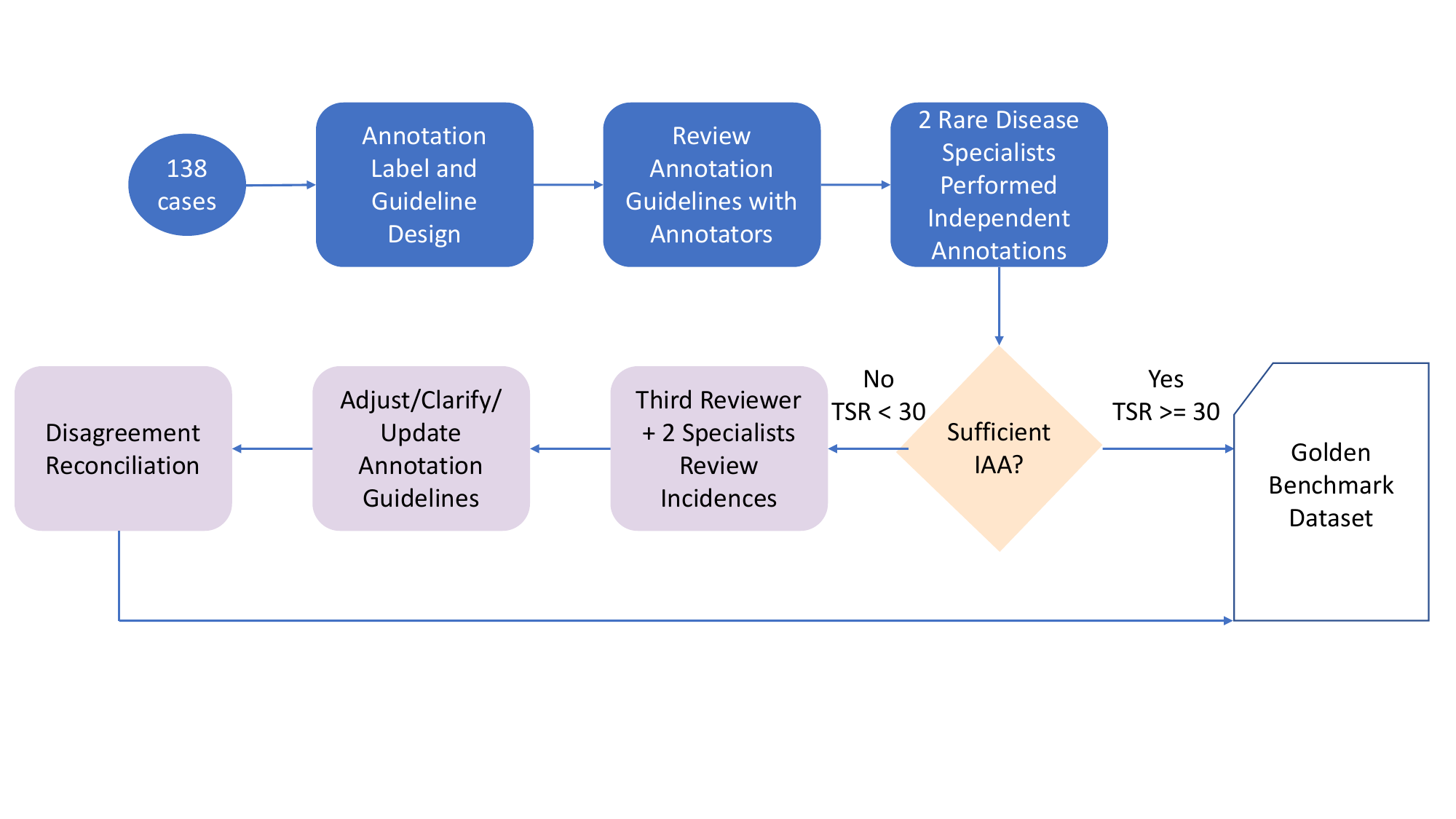}
    \caption{Annotation and Consolidation Pipeline.}
    \label{fig:annotation_workflow}
\end{figure*}

\section{Category-Wise Annotation Comparison Between Physicians}
This appendix presents a comparison of the initial category-wise annotations made by two rare disease specialists.
\label{tab:annotation_metrics}

\begin{figure*}
    \centering
    \includegraphics[width=\textwidth]{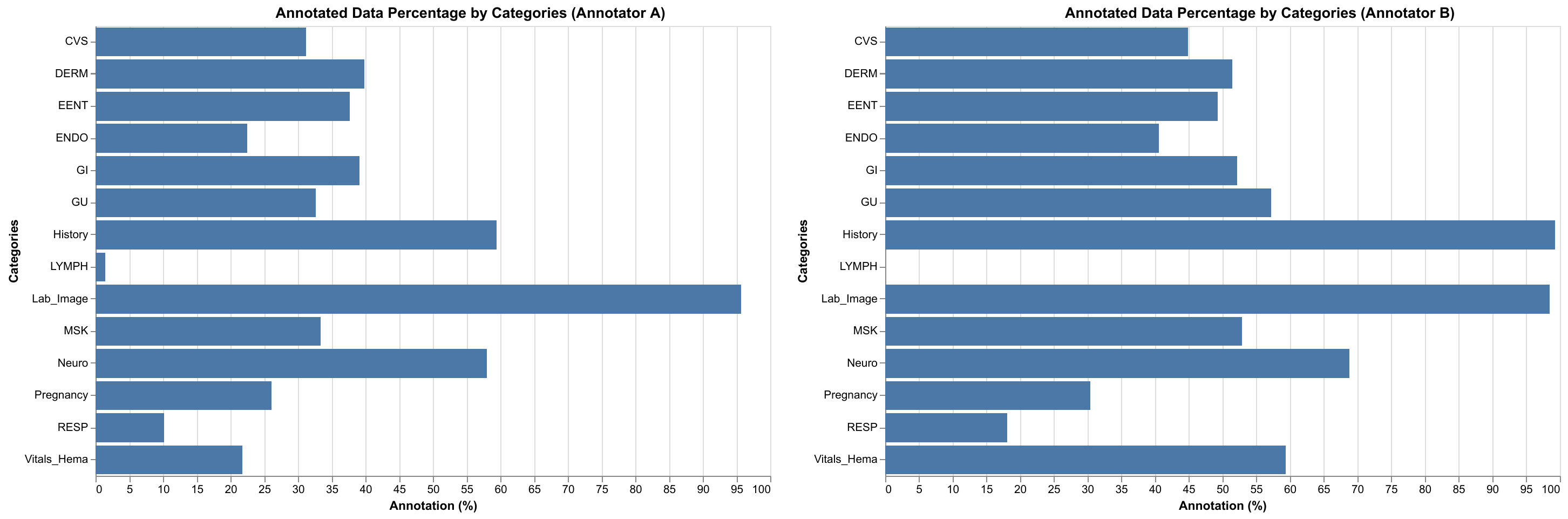}
    \caption{Category-Wise Annotation Comparison Between CF and FR}
    \label{fig:annotation-comparison}
\end{figure*}

\textbf{Figure~\ref{fig:annotation-comparison}} illustrates the percentage of data annotated by each annotator for various medical categories. The x-axis represents the percentage of annotations, while the y-axis lists the categories. Notable differences are observed in categories such as \textit{Lab\_Image} and \textit{History}, where physician B appears to annotate a larger proportion compared to physician A.

\textbf{Table~\ref{tab:across_all_metrics}} provides a quantitative evaluation of intra-annotator agreement using three metrics:
- \textbf{Average TSR(\%)(Token Set Ratio)}: Measures the similarity of token sets between annotations, normalized for the longer text length.
- \textbf{Average Levenshtein}: Quantifies the minimum edit distance required to make one annotation identical to another, normalized for length.
- \textbf{Average Exact Match}: Indicates the proportion of annotations that are identical without any modifications.

These metrics demonstrate high agreement in certain categories such as \textit{LYMPH} and \textit{RESP}, while lower agreement is observed in categories like \textit{History} and \textit{Lab\_Image}. This indicates potential differences in annotation styles or understanding of these specific categories.

\begin{table*}[h]
\centering 
\begin{tabular}{@{}lccc@{}}
\toprule
\textbf{Category} & \textbf{Average TSR(\%)} & \textbf{Average Levenshtein} & \textbf{Average Exact Match} \\ 
\midrule
CVS           & 78.434 & 0.719 & 0.587 \\ 
DERM          & 80.753 & 0.767 & 0.500 \\ 
EENT          & 82.539 & 0.746 & 0.507 \\ 
ENDO          & 70.322 & 0.662 & 0.565 \\ 
GI            & 79.693 & 0.683 & 0.493 \\ 
GU            & 71.820 & 0.662 & 0.449 \\ 
History       & 42.501 & 0.292 & 0.007 \\ 
LYMPH         & 98.551 & 0.986 & 0.986 \\ 
Lab\_Image    & 54.870 & 0.723 & 0.036 \\ 
MSK           & 67.748 & 0.615 & 0.457 \\ 
Neuro         & 72.126 & 0.699 & 0.304 \\ 
Pregnancy     & 93.047 & 0.867 & 0.703 \\ 
RESP          & 91.329 & 0.873 & 0.819 \\ 
Vitals\_Hema  & 61.367 & 0.582 & 0.428 \\ 
\textbf{Average} & \textbf{74.650} & \textbf{0.705} & \textbf{0.489} \\ 
\bottomrule
\end{tabular}
\caption{Category-Wise Comparison of Annotations}
\label{tab:across_all_metrics}
\end{table*}

\clearpage

\section{Inter-Annotation Agreement for Initial Expert Annotations Before Reconciliation}
The below statistics were auto-generated by Prodigy annotation pipeline
\label{appendix:iaa}

\subsection*{Annotation Statistics}
The annotation process involved using 14 label categories and two clinical specialists. Table \ref{tab:annotation_statistics} provides a summary of the annotation statistics, including the number of examples, categories, and co-incident examples.

\begin{table}[h!]
\centering
\caption{Annotation Statistics}
\label{tab:annotation_statistics}
\begin{tabular}{ll}
\hline
\textbf{Attribute}                 & \textbf{Value} \\ \hline
Examples                           & 276            \\
Categories                         & 14             \\
Co-Incident Examples*              & 138            \\
Single Annotation Examples         & 0              \\
Annotators                         & 2              \\
Avg. Annotations per Example       & 2.0            \\ \hline
\end{tabular}
\end{table}
\noindent \textit{*Co-incident examples: Examples with more than one annotation.}

\subsection*{Agreement Statistics}
Table \ref{tab:agreement_statistics} summarizes the agreement statistics between annotators using pairwise F1 scores for each category. The ``Support" column indicates the number of instances for each category.

\begin{landscape}
\begin{table}[htbp]
\centering
\begin{tabular}{@{}lcc@{}}
\toprule
\textbf{Category}      & \textbf{Pairwise F1} & \textbf{Support} \\ \midrule
Vitals\_Hema           & 0.33                & 40               \\
RESP                   & 0.36                & 34               \\
Lab\_Image             & 0.55                & 400              \\
LYMPH                  & 0.0                 & 2                \\
DERM                   & 0.56                & 136              \\
History                & 0.57                & 148              \\
EENT                   & 0.53                & 88               \\
Neuro                  & 0.51                & 245              \\
Pregnancy              & 0.54                & 44               \\
GI                     & 0.42                & 103              \\
CVS                    & 0.50                & 81               \\
MSK                    & 0.36                & 82               \\
GU                     & 0.33                & 79               \\
ENDO                   & 0.18                & 38               \\ \bottomrule
\end{tabular}
\caption{Agreement statistics between annotators for each category, using pairwise F1 scores. The ``Support" column shows the number of instances in each category.}
\label{tab:agreement_statistics}
\end{table}
\end{landscape}

\subsection*{Confusion Matrix}
This section presents the \textbf{confusion matrix} derived from the initial annotations provided by two medical experts before the reconciliation process. Each row represents a \textbf{predicted category}, while each column represents the \textbf{actual category} assigned in the benchmark. The values indicate the proportion of instances classified into each category, highlighting areas of \textbf{agreement} and \textbf{misclassification} between experts.

The matrix reveals several important trends in expert annotations. Most categories exhibit strong \textbf{self-consistency}, as indicated by high diagonal values. Categories such as \textbf{Vitals\_Hema (0.98), EENT (0.92), Pregnancy (0.91), and DERM (0.91)} demonstrate excellent agreement between experts, suggesting that these categories are well-defined and consistently understood. The \textbf{LYMPH} category shows perfect agreement (\textbf{1.0}), indicating no cases of misclassification between experts in this category.

Despite these high agreement rates, some categories exhibit substantial \textbf{misclassification}, particularly in cases where medical domains overlap. \textbf{Lab\_Image} cases are occasionally misclassified into \textbf{DERM (0.01), Neuro (0.01), CVS (0.01), and NONE (0.2)}, suggesting difficulties in differentiating between diagnostic information across different systems. Similarly, \textbf{RESP } cases show occasional misclassification into \textbf{GI (0.03)}, likely due to the presence of symptoms that can be associated with both systems, such as coughing. Additionally, \textbf{MSK} and \textbf{Neuro} categories exhibit some level of confusion, which may be attributed to the overlap between nerve-related pain and movement disorders.

One of the most challenging categories is \textbf{NONE}, which exhibits substantial misclassification. Many instances classified as \textbf{NONE (0.28)} spill into \textbf{Lab\_Image (0.11), Neuro (0.12), and ENDO (0.07)}, indicating that some cases initially deemed unclassifiable were later reassigned to specific medical domains during reconciliation. Similarly, broad categories such as \textbf{History, Lab\_Image, and NONE} exhibit higher misclassification rates, suggesting that they encompass complex multi-system cases that were more prone to expert disagreement.

The initial expert annotations highlight both areas of \textbf{strong agreement} and cases of \textbf{ambiguity} where refinement may be necessary. The high agreement within well-defined categories demonstrates the effectiveness of existing classification criteria, whereas the observed misclassification trends underscore the challenges in distinguishing between closely related medical domains. These findings provide a foundation for further refinement of annotation practices and improved classification approaches in \textbf{dense medical information extraction}.

\begin{table*}
\centering
\resizebox{\linewidth}{!}{
\begin{tabular}{|l|c|c|c|c|c|c|c|c|c|c|c|c|c|c|c|}
\hline
            & Vitals\_Hema & RESP & Lab\_Image & LYMPH & DERM & History & EENT & Neuro & Pregnancy & GI & CVS & MSK & GU & ENDO & NONE \\ \hline
Vitals\_Hema & 0.98 & 0.0  & 0.0  & 0.0  & 0.0  & 0.0  & 0.0  & 0.0  & 0.0  & 0.0  & 0.0  & 0.0  & 0.0  & 0.0  & 0.02 \\ \hline
RESP         & 0.0  & 0.59 & 0.0  & 0.0  & 0.0  & 0.0  & 0.0  & 0.0  & 0.03 & 0.03 & 0.03 & 0.0  & 0.0  & 0.0  & 0.32 \\ \hline
Lab\_Image   & 0.0  & 0.0  & 0.74 & 0.0  & 0.01 & 0.0  & 0.0  & 0.01 & 0.0  & 0.0  & 0.01 & 0.01 & 0.01 & 0.0  & 0.2  \\ \hline
LYMPH        & 0.0  & 0.0  & 0.0  & 0.0  & 0.0  & 0.0  & 0.0  & 0.0  & 0.0  & 0.0  & 0.0  & 0.0  & 0.0  & 0.0  & 1.0  \\ \hline
DERM         & 0.0  & 0.0  & 0.0  & 0.0  & 0.91 & 0.0  & 0.0  & 0.0  & 0.0  & 0.0  & 0.0  & 0.04 & 0.01 & 0.0  & 0.04 \\ \hline
History      & 0.0  & 0.0  & 0.01 & 0.0  & 0.0  & 0.77 & 0.03 & 0.01 & 0.01 & 0.0  & 0.01 & 0.01 & 0.0  & 0.0  & 0.16 \\ \hline
EENT         & 0.0  & 0.0  & 0.0  & 0.0  & 0.0  & 0.0  & 0.92 & 0.01 & 0.0  & 0.0  & 0.0  & 0.0  & 0.0  & 0.0  & 0.07 \\ \hline
Neuro        & 0.0  & 0.0  & 0.0  & 0.0  & 0.0  & 0.0  & 0.0  & 0.89 & 0.0  & 0.0  & 0.0  & 0.03 & 0.0  & 0.0  & 0.07 \\ \hline
Pregnancy    & 0.0  & 0.0  & 0.0  & 0.0  & 0.0  & 0.0  & 0.0  & 0.0  & 0.91 & 0.0  & 0.0  & 0.0  & 0.0  & 0.0  & 0.09 \\ \hline
GI           & 0.01 & 0.0  & 0.0  & 0.0  & 0.0  & 0.0  & 0.01 & 0.0  & 0.0  & 0.85 & 0.0  & 0.0  & 0.0  & 0.0  & 0.13 \\ \hline
CVS          & 0.0  & 0.0  & 0.01 & 0.0  & 0.0  & 0.01 & 0.0  & 0.01 & 0.0  & 0.0  & 0.8  & 0.0  & 0.0  & 0.0  & 0.16 \\ \hline
MSK          & 0.0  & 0.0  & 0.0  & 0.0  & 0.0  & 0.01 & 0.0  & 0.01 & 0.0  & 0.0  & 0.0  & 0.73 & 0.0  & 0.01 & 0.23 \\ \hline
GU           & 0.03 & 0.0  & 0.0  & 0.0  & 0.0  & 0.01 & 0.0  & 0.0  & 0.0  & 0.0  & 0.0  & 0.0  & 0.77 & 0.03 & 0.16 \\ \hline
ENDO         & 0.0  & 0.0  & 0.0  & 0.0  & 0.0  & 0.0  & 0.0  & 0.0  & 0.0  & 0.0  & 0.0  & 0.0  & 0.0  & 0.68 & 0.32 \\ \hline
NONE         & 0.05 & 0.02 & 0.11 & 0.0  & 0.05 & 0.04 & 0.04 & 0.12 & 0.02 & 0.07 & 0.03 & 0.05 & 0.07 & 0.07 & 0.28 \\ \hline
\end{tabular}
}
\caption{Confusion Matrix}
\label{tab:confusion_matrix}

\end{table*}

\end{document}